\documentclass[10pt,twocolumn,letterpaper]{article}

\usepackage{cvpr}              
\definecolor{cvprblue}{rgb}{0.21,0.49,0.74}
\usepackage[pagebackref,breaklinks,colorlinks,allcolors=cvprblue]{hyperref}

\usepackage{booktabs}       
\usepackage{array}
\usepackage{makecell}
\usepackage{multirow}
\usepackage{colortbl}
\usepackage{xcolor}   
\usepackage{caption} 
\usepackage{stfloats}
\usepackage{float}
\usepackage{fontawesome}
\usepackage{amsmath}   
\usepackage{amssymb}   
\usepackage{amsfonts}  
\usepackage{tabularray}
\usepackage{pifont}

\definecolor{darkgrey}{rgb}{0.25, 0.25, 0.25} 

\def\paperID{7318} 
\def\confName{CVPR}
\def\confYear{2026}

\title{LoD-Loc v3: Generalized Aerial Localization in Dense Cities using Instance Silhouette Alignment}

\author{Shuaibang Peng$^{1}$, Juelin Zhu$^{1\dagger}$, Xia Li$^{1}$, Kun Yang$^2$, Maojun Zhang$^{1}$, Yu Liu$^{1}$, Shen Yan$^{1\dagger}$
\and
$^1$National University of Defense Technology, \ $^2$Northwestern Polytechnical University\\
{\tt\small \{psb24,zhujuelin,lixia24,mjzhang,jasonyuliu,yanshen12\}@nudt.edu.cn
}
}

\begin{document}

\twocolumn[{%
    \renewcommand\twocolumn[1][]{#1}
    \maketitle 

    \begin{center}
        \centering
        \includegraphics[width=\textwidth]{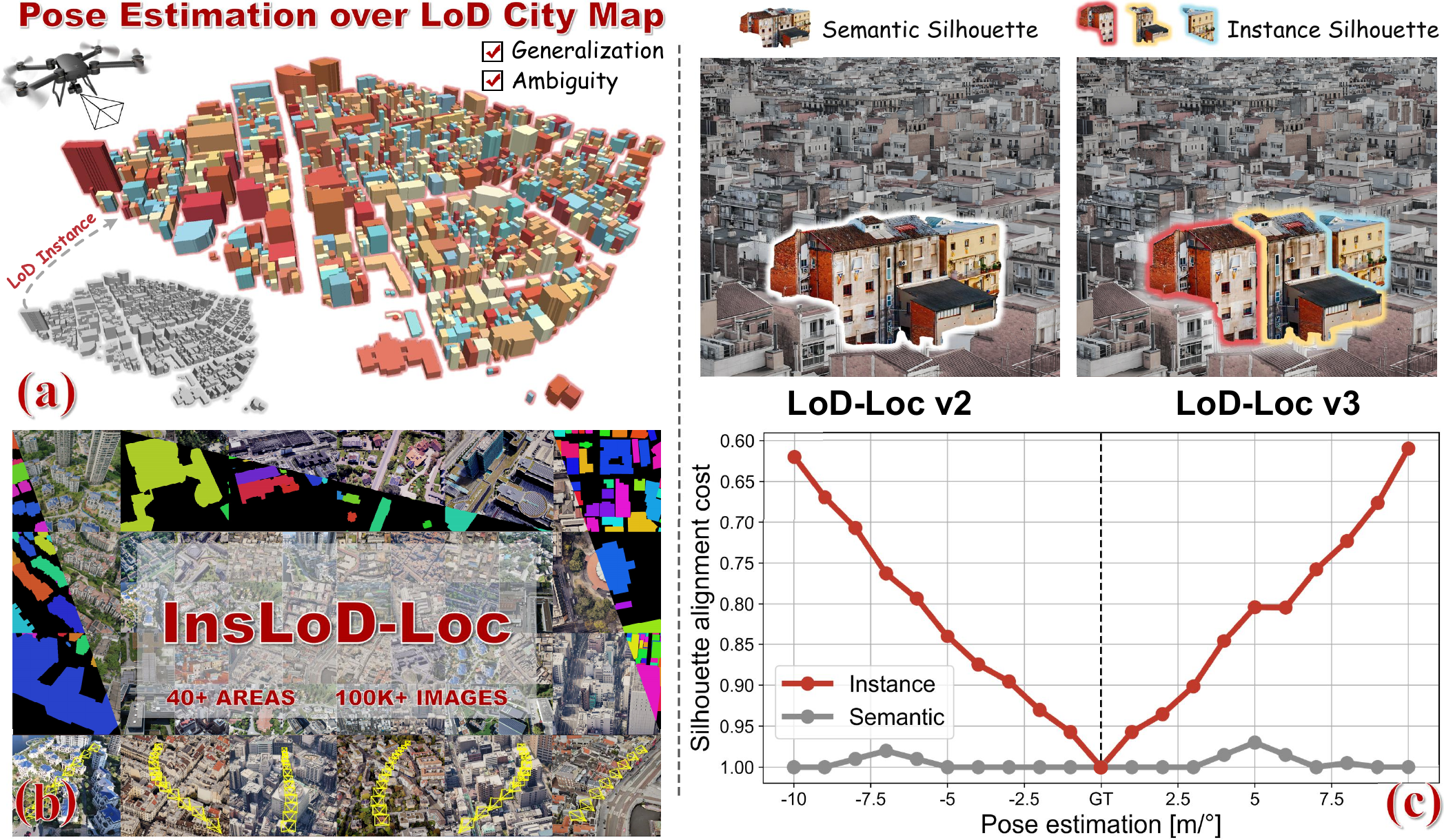}
        \captionof{figure}{In this paper, (a) we introduce LoD-Loc v3 to address two critical challenges in aerial localization over LoD city models: cross-scene generalization and the ambiguity problem in dense urban scenes. Our solutions are twofold: (b) we construct \textbf{InsLoD -Loc}, a large-scale synthetic dataset covering 40 distinct areas for model zero-shot training, and (c) we reformulate the localization paradigm by shifting from semantic to instance silhouette alignment, which provides superior convergence.}
        \label{fig:highlight}
    \end{center}%
}]
\renewcommand{\thefootnote}{$\dagger$}
\footnotetext{Corresponding author}
\begin{abstract}
We present LoD-Loc v3, a novel method for generalized aerial visual localization in dense urban environments.
While prior work LoD-Loc v2~\cite{zhu2025lod} achieves localization through semantic building silhouette alignment with low-detail city models, it suffers from two key limitations: poor cross-scene generalization and frequent failure in dense building scenes. 
Our method addresses these challenges through two key innovations. First, we develop a new synthetic data generation pipeline that produces \textbf{InsLoD-Loc} - the largest instance segmentation dataset for aerial imagery to date, comprising 100k images with precise instance building annotations. This enables trained models to exhibit remarkable zero-shot generalization capability. Second, we reformulate the localization paradigm by shifting from semantic to instance silhouette alignment, which significantly reduces pose estimation ambiguity in dense scenes.
Extensive experiments demonstrate that LoD-Loc v3 outperforms existing state-of-the-art (SOTA) baselines, achieving superior performance in both cross-scene and dense urban scenarios with a large margin.
The project is available at {\definecolor{myurlcolor}{HTML}{ED028C}\hypersetup{urlcolor=myurlcolor}\url{https://nudt-sawlab.github.io/LoD-Locv3/}}.
\end{abstract}     
\section{Introduction}
\label{sec:intro}

\noindent Visual localization for unmanned aerial vehicle (UAV) estimates a camera's pose by matching visual inputs against a georeferenced map, which is a core technology for autonomous systems, supporting applications such as precision navigation~\cite{lu2018survey}, cargo transport~\cite{villa2020survey}, and emergency response~\cite{cai2022review, wu2024uavd4l}. The dominant paradigm~\cite{chen2021real, wu2024uavd4l, yan2023render} relies on high-fidelity 3D reconstructions, typically generated using Structure-from-Motion (SfM)~\cite{schonberger2016structure} or photogrammetry~\cite{panek2022meshloc}. While effective, these high-fidelity models are expensive to create and maintain, and their large data volume raises privacy and protection concerns.

To address these problems, localization methods based on Level-of-Detail (LoD) city models have emerged as a promising alternative~\cite{zhu2024lod,zhu2025lod}. The LoD model is a 3D representation focused on architectural structures, following the CityGML standard~\cite{groger2012ogc, kutzner2023ogc, wysocki2024reviewing}.
The construction of LoD models has become a global trend, with nationwide productions in the USA~\cite{open_city_model}, China~\cite{china_LoD1, SUN2024114057}, Switzerland~\cite{swisstopo, Swiss-LoD1}, Japan~\cite{Japan_LoD1}, Singapore~\cite{SG-LoD1, SG-LoD2}, Germany~\cite{Das_3D}, and the Netherlands~\cite{3D_BAGS, Netherland-LoD1}.
Academia and industry are also accelerating this, with open-source global datasets~\cite{zhu2025globalbuildingatlas} and AI-driven commercial mapping~\cite{Vividfeatures}. 
As a result, using LoD models for localization has the potential to support global UAV navigation in urban environments.

The early LoD-Loc~\cite{zhu2024lod} pioneered this direction by aligning wireframes from high-detail LoD models. LoD-Loc v2~\cite{zhu2025lod} extended this to ubiquitous low-detail LoD1 models by aligning building silhouettes. However, LoD-Loc v2~\cite{zhu2025lod} suffers from two critical drawbacks: it usually fails to generalize to unseen scenes, and it completely fails to localize in dense scenes because of the ambiguity of the semantic silhouette, as illustrated in~\cref{fig:highlight}.

To overcome the generalization and ambiguity challenges, we propose LoD-Loc v3. For generalization, we developed a novel synthetic data pipeline to construct \textbf{InsLoD-Loc}, a 100k-image dataset with precise instance-level building annotation. To our knowledge, this is the largest aerial instance segmentation dataset, and our pipeline is designed for easy extension. Specifically, this pipeline involves two stages: First, we render the photorealistic RGB images within the Unreal Engine 5 (UE5)~\cite{UE_engine} platform, which uses the Cesium for Unreal plugin~\cite{Cesium_GS} to stream Google Earth Photorealistic 3D Tileset data~\cite{GoogleEarth} and the AirSim plugin~\cite{Microsoft_Research} for capture. Second, we source the corresponding LoD models from~\cite{zhu2025lod, swisstopo, 3D_BAGS, PLATEAU, zamir2016generic}, align their coordinate systems with the 3D tileset data, and employ OpenSceneGraph (OSG)~\cite{OSG} with our model instancing method (\cref{sec:scaling_to_open_world}) to render precise instance masks using the identical camera poses with RGB-based rendering.

To address the ambiguity problem, we shift the core paradigm from semantic silhouettes to instance silhouettes alignment. Our motivation is that localization in cities should be viewed as an instance alignment process, which, as shown in~\cref{fig:highlight}, is particularly critical in dense environments. Specifically, our method involves three stages: First, a novel 'LoD model instancing' preprocessing step assigns unique identifiers to each building for instance-aware rendering. Second, we use our constructed \textbf{InsLoD-Loc} dataset to fine-tune a SAM-based model to extract building instance silhouettes from the query image. Finally, we estimate the pose by aligning the query and rendered instance silhouettes, similar to the coarse-to-fine localization framework in LoD-Loc v2~\cite{zhu2025lod}.

On public benchmarks, our method exceeds current SOTA baselines, notably without being trained in-distribution. Besides, we constructed a dataset containing multiple dense urban scenes. On this challenging dataset, our localization results demonstrate a \textbf{2000\%} improvement over SOTA methods at the $(2\text{m}, 2^{\circ})$ accuracy.

In summary, our main contributions are as follows:
\begin{enumerate}

    \item We propose a synthetic data pipeline and construct a large-scale dataset, \textbf{InsLoD-Loc}, which addresses poor cross-scene generalization.
    \item We introduce a novel localization method based on instance silhouettes, which resolves the ambiguity problem in dense scenes.
    \item We demonstrate through extensive experiments that our method achieves SOTA performance on two public benchmarks and our own dense-scene dataset.
\end{enumerate}

\section{Related Works}
\label{sec:Related_Works}

\noindent \textbf{Localization over 3D Models.}
\noindent Mainstream high-precision visual localization~\cite{wu2024uavd4l, sun2021loftr, sarlin2020superglue, sarlin2019coarse, panek2022meshloc, cheng2026pilotneuralpixelto3dregistration} typically relies on information-rich 3D models like SfM point clouds~\cite{schonberger2016structure} or textured meshes~\cite{panek2022meshloc}. The core of these methods is to establish 2D-to-3D correspondences, often involving image retrieval~\cite{arandjelovic2016netvlad,ge2020self,fu2025adaptive} followed by local feature matching~\cite{lowe2004distinctive,detone2018superpoint,karkus2018particle,sarlin2020superglue,sun2021loftr}. A 6-DoF pose is then solved via a PnP-RANSAC algorithm~\cite{haralick1994review,kneip2011novel,barath2019magsac,chum2008optimal,fischler1981random,chum2003locally,yan2023long,lepetit2009ep}. While accurate, these methods depend on detailed maps that are costly to create and maintain, computationally intensive, raise privacy and protection concerns.

\noindent \textbf{Localization over LoD Models.}
Early LoD localization was limited to ground-level views~\cite{ramalingam2010skyline2gps,ramalingam2011pose, saurer2016image}. LoD-Loc~\cite{zhu2024lod} pioneered aerial localization by aligning neural wireframes with high-detail LoD2/3 models. LoD-Loc v2~\cite{zhu2025lod} adapted this to more common low-detail LoD1 models by aligning building silhouettes. However, its reliance on semantic segmentation led to two fundamental challenges: poor generalization due to limited training data, and catastrophic localization failure in dense urban scenes because of ambiguity. 
To address these, LoD-Loc v3 enhances generalization and ambiguity by introducing a large-scale synthetic dataset and shifting the paradigm from semantic to instance-level alignment.

\noindent \textbf{Instance Segmentation.}
The goal of instance segmentation is to identify pixel-level and instance-level silhouettes for every object in an image. Traditional methods are two-stage~\cite{he2017mask,chen2019hybrid,wang2019panet,huang2019mask,kirillov2020pointrend} or single-stage~\cite{bolya2019yolact,chen2019tensormask,wang2020solov2,chen2020blendmask,lee2020centermask}. Vision foundation models like SAM~\cite{kirillov2023segment} introduced a paradigm shift. Their zero-shot capabilities spurred follow-up works on efficiency~\cite{zhao2023fast}, quality~\cite{ke2024segment}, and domain adaptation~\cite{chen2024rsprompter,chen2023sam,chen2024spectral,chen2026local}. However, the success of these data-driven methods is highly dependent on large-scale, high-quality annotated datasets. While many benchmarks exist for general~\cite{lin2015microsoft,gupta2019lvis,shao2019objects365}, medical~\cite{sirinukunwattana2015stochastic,jha2020kvasir}, or remote sensing~\cite{waqas2019isaid,su2019object} domains, they have a significant domain gap with our task. Even the WHU Building dataset~\cite{ji2018fully} consists of nadir-view satellite imagery, unlike the low-altitude, oblique UAV views in urban canyons. 
While previous works demonstrated the value of synthetic training data for vision tasks~\cite{mayer2018makes} and the feasibility of simulation for feature evaluation, multi-object system identification, and robotic control~\cite{gao2020local,agarwal2023simulating,liumulti,song2025humeintroducingsystem2thinking}, there remains a lack of dedicated instance-level building datasets for localization.
To fill this gap, we constructed our 100k-image synthetic dataset to support research in building instance segmentation and localization from low-altitude aerial perspectives.
\section{Method}
\label{sec:Method}

\noindent \textbf{Problem Formulation.}
Given a LoD model $\mathcal{M}$, a query image $I_q$, and its prior pose estimate ${\boldsymbol{\xi}}^p_q$, the objective is to estimate the absolute pose $\boldsymbol{\xi}^*_q$ of $I_q$.

\noindent \textbf{Overview.} 
In the following sections, we first briefly review the localization framework of LoD-Loc v2~\cite{zhu2025lod} as a preliminary. We then introduce our core contributions: a large-scale dataset constructed via our proposed synthetic data generation pipeline (detailed in~\cref{sec:Data_Generation}) to address the problem of poor cross-scene generalization, and a localization algorithm based on instance silhouette alignment to resolve the issue of ambiguity in dense scenes.

\subsection{The Coarse-to-Fine Localization Framework}
\label{sec:coarse_to_fine_framework}

\noindent UAV inertial sensors reliably provide gravity direction, reducing the problem to a 4-DoF pose ${\boldsymbol{\xi}}^*_q = (x, y, z, \theta)$ (3D translation, 1D yaw)~\cite{yan2023long, sarlin2023orienternet, sarlin2024snap, lynen2020large, Fragoso_2020_CVPR}. LoD-Loc v2~\cite{zhu2025lod} solves this by aligning a rendered silhouette $M_{hyp}$ from the LoD model $\mathcal{M}$ with a predicted image silhouette $M_q$, quantified by Intersection over Union (IoU):
\begin{equation}
    \label{eq:cost_v2}
    \mathrm{c}(I_q, I_{hyp}) = \frac{|M_q \cap M_{hyp}|}{|M_q \cup M_{hyp}|}
\end{equation}

\noindent The \textbf{coarse stage} finds an initial pose ${\boldsymbol{\xi}}^c_q$ by uniformly sampling the 4-DoF search space $(x, y, z, \theta)$ and maximizing $\mathrm{c}$. The \textbf{fine stage} then uses a particle filter to refine this estimate, iteratively maximizing the cost $\mathrm{c}$ to find the final pose ${\boldsymbol{\xi}}^*_q$~\cite{karkus2018particle, lin2023parallel, maggio2023loc, humenberger2022investigating, kendall2015posenet, trivigno2024unreasonable, Pietrantoni_2023_CVPR}. Detailed descriptions of both stages can be found in~\cite{zhu2025lod}.

\subsection{Scaling to Open-World Scenes}
\label{sec:scaling_to_open_world}

\noindent Although LoD-Loc v2~\cite{zhu2025lod} performs well in specific scenes, its performance degrades significantly or fails entirely in cross-scene or dense urban environments. To address these issues, we propose a new methodology designed to enhance model generalization and ambiguity. 

To improve generalization, we constructed \textbf{InsLoD-Loc}, a large-scale synthetic dataset (\cref{sec:Dataset}). To address ambiguity in dense scenes, we replace the semantic silhouette cost (\cref{eq:cost_v2}) with an instance silhouette alignment.

Our approach has three stages: First, an instance method for LoD models, assigning a unique ID to each building. Second, an instance segmentation model to extract individual building silhouettes from the query image. Finally, a novel alignment and evaluation mechanism that leverages these instance silhouettes.

\begin{figure}[t]
    \centering
	\includegraphics[ width=1.0\linewidth]{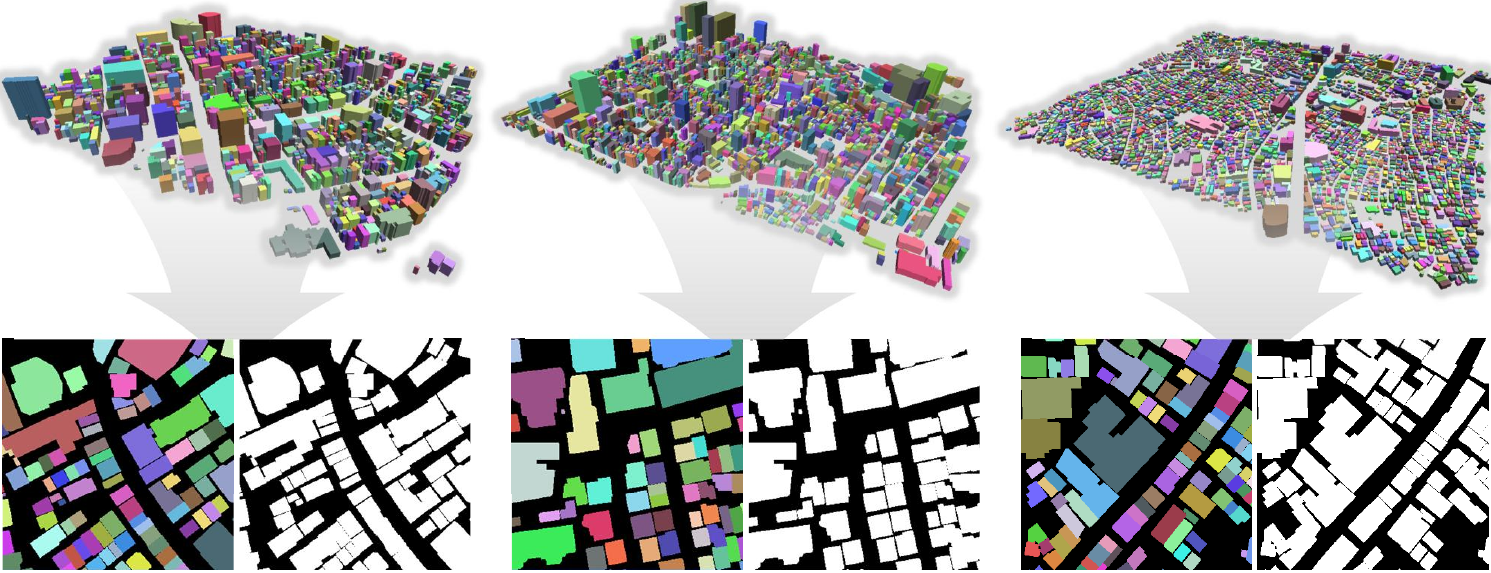}
         \caption{\textbf{Visualization of instanced LoD models and labels.} The top row displays the instanced LoD models. The bottom row illustrates the rendered instance labels rendered from the models, alongside their corresponding semantic labels.}
    \label{fig:instance_obj}
\end{figure}

\noindent \textbf{LoD Model Instancing.} 
We assign a unique color to each building in $\mathcal{M}$ via topological structure analysis. By parsing the untextured $\mathcal{M}$ as a graph $G = (V, E, F)$, each building $B_i$ forms a connected component $G_i$, formulating instancing as a graph partition:
\begin{equation}
    \label{eq:graph_partition}
    \mathcal{M} = \bigcup_{i=1}^{M} B_i, \quad \text{s.t.} \quad B_i \cap B_j = \emptyset \quad \forall i \neq j
\end{equation}
where $M$ is the number of buildings. Each $B_i$ receives a unique 24-bit integer ID mapped to an RGB color (details in Appendix). 
Assigning this color to all faces of $B_i$ yields the instanced LoD model $\mathcal{M}_{ins}$ as in \cref{fig:instance_obj}.
Rendering $\mathcal{M}_{ins}$ from any pose $\boldsymbol{\xi}_{hyp}$ produces an Instance Map $I_{hyp}$, where colors uniquely encode building identities.

\noindent \textbf{Building Silhouette Instance Segmentation.} This stage segments each building silhouette from $I_q$. We fine-tune a SAM-based model~\cite{kirillov2023segment} on our \textbf{InsLoD-Loc} dataset, inspired by prompt learning~\cite{chen2024rsprompter}. The network, composed of a SAM image encoder, a learnable Prompter Module, and a SAM mask decoder, is adapted to extract precise building silhouettes from aerial images. Specifically, for a given input query image \( I_q \in \mathbb{R}^{3 \times H \times W} \), the SAM encoder extracts an image embedding $F_{embed}$. The Prompter Module predicts prompt embeddings from $F_{embed}$. The SAM decoder takes both embeddings to generate instance masks $\mathcal{S}_q = \{M_q^j\}_{j=1}^{N}$. This transforms SAM into an automatic, task-specific segmentation pipeline, providing unambiguous geometric constraints and resolving ambiguity from merged silhouettes. Further details are in the Appendix.

\noindent \textbf{Pose Evaluation via Instance Alignment.} The core of our localization framework is a novel cost function, $\mathrm{c}_{ins}$, used to evaluate any given pose hypothesis $\boldsymbol{\xi}_{hyp}$. For each hypothesis, we first project the instanced LoD model $\mathcal{M}_{ins}$ using OSG~\cite{OSG} real-time rendering to obtain its set of instance masks, $\mathcal{S}_{hyp} = \{M_{hyp}^k\}_{k=1}^{K}$. We then compute the alignment cost between this set and the query instance set $\mathcal{S}_q = \{M_q^j\}_{j=1}^{N}$ using an asymmetric matching strategy. For each predicted instance $M_q^j$, we find its best match in $\mathcal{S}_{hyp}$ (highest Dice coefficient) $d_j^*$:
\begin{equation}
    \label{eq:best_match_dice}
    d_j^* = \max_{k \in \{1, \dots, K\}} \left( \frac{2 \sum_{p} (M_q^j(p) \cdot M_{hyp}^k(p))}{\sum_{p} M_q^j(p) + \sum_{p} M_{hyp}^k(p) + \epsilon} \right)
\end{equation}
where $p$ is a pixel location, and $\epsilon$ is a small constant to prevent division by zero. The final cost $\mathrm{c}_{ins}$ is a weighted sum of $d_j^*$. We explore two weighting strategies:

\begin{itemize}
    \item \textbf{Confidence-based Weighting:} This strategy uses the instance's confidence score $s_j$ as weight $w_j^{(\text{conf})}$.
    \begin{equation}
        \label{eq:instance_cost_conf}
        \mathrm{c}_{ins}^{(\text{conf})} = \sum_{j=1}^{N} \frac{s_j}{\sum_{i=1}^{N} s_i} \cdot d_j^*
    \end{equation}

    \item \textbf{Area-based Weighting:} This strategy uses the instance's bounding box area $A_j$ as weight $w_j^{(\text{area})}$.
    \begin{equation}
        \label{eq:instance_cost_area}
        \mathrm{c}_{ins}^{(\text{area})} = \sum_{j=1}^{N} \frac{A_j}{\sum_{i=1}^{N} A_i} \cdot d_j^*
    \end{equation}
\end{itemize}
We adopt the coarse-to-fine localization framework from LoD-Loc v2~\cite{zhu2025lod}, using $\mathrm{c}_{ins}$ as the core metric. In the coarse stage, we construct the cost volume $\mathcal{C}_{Ins}$ by computing this value over a uniformly sampled grid to select an initial coarse pose $\boldsymbol{\xi}^c_q$. Subsequently, in the fine stage, this cost is used to evaluate particle weights, guiding the optimization process to converge to the final, precise pose $\boldsymbol{\xi}^*_q$.

\noindent \textbf{Supervision.} Our instance segmentation network is trained in a supervised manner using pairs of query images $I_q$, ground truth masks $G$, and corresponding bounding boxes $B$. During training, we update the learnable Prompter Module and the SAM mask decoder, while applying LoRA~\cite{hu2022lora} for parameter-efficient fine-tuning to the otherwise frozen SAM vision encoder, significantly reducing computational overhead. The model's overall loss function $L$ is a multi-task loss, composed of a Region Proposal Network (RPN) loss $L_{\text{rpn}}$ and a Region of Interest (RoI) head loss $L_{\text{roi}}$:
\begin{equation}
    L = L_{\text{rpn}} + L_{\text{roi}}
    \label{eq:total_loss}
\end{equation}
where $L_{\text{rpn}}$ supervises candidate box generation within the Prompter Module, and $L_{\text{roi}}$ supervises the final classification, regression, and the SAM decoder's mask predictions.

\begin{figure*}[htbp]
  \centering 
  \includegraphics[width=1.0\linewidth]{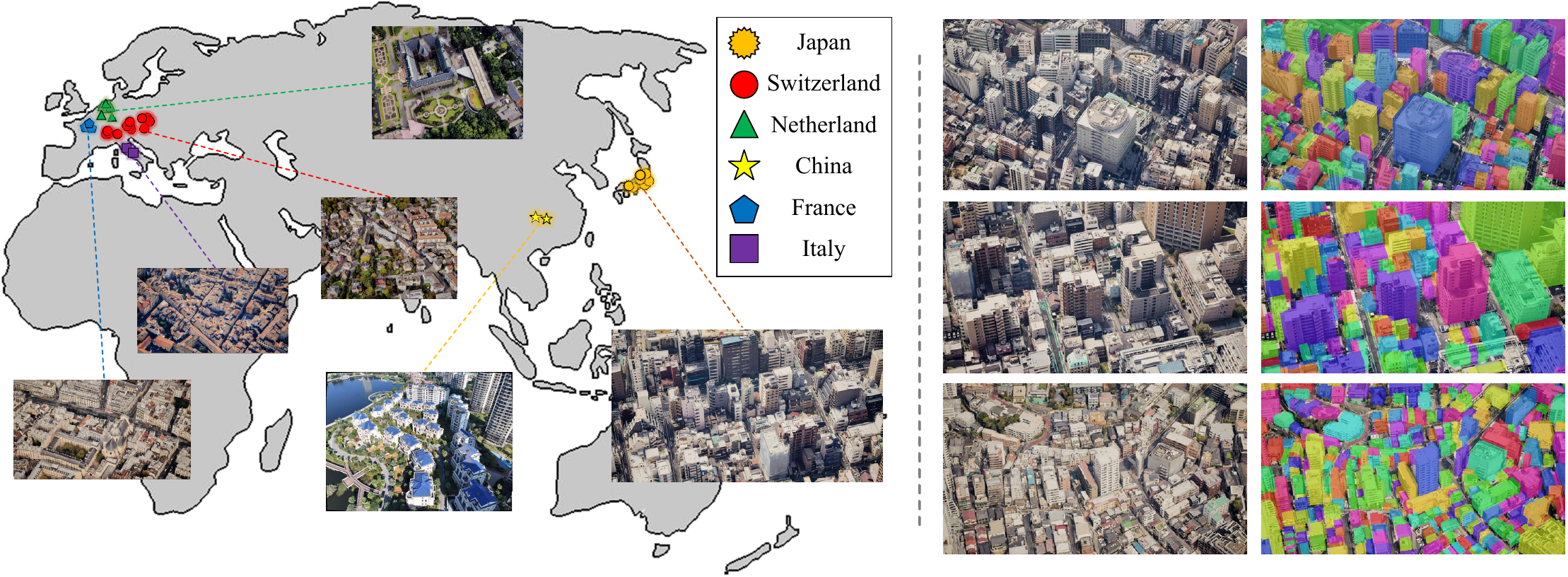}  
  \caption{\textbf{Overview of the InsLoD-Loc dataset.} The left panel illustrates the geographic distribution of the 40 flight areas across Europe and Asia. The right panel showcases representative samples from the dataset, each displaying (from left to right): a photorealistic RGB query image and its corresponding pixel-accurate instance label, where each color represents a unique building.}
  \label{fig:data}    
\end{figure*}
\begin{table*}[htbp]
\centering
\begin{tabular}{cccccccc} 
\toprule

\multirow{2}{*}{{\textbf{Camera}}} & \multicolumn{3}{c}{{\textbf{Camera Configuration}}}   & \multicolumn{3}{c}{{\textbf{Sampling Methodology}}}          \\ 
\cmidrule(r){2-4} \cmidrule(r){5-8}
                        & Resolution[px] & Sensor Size          & FOV & Strategy         & Path       & View Angle & Height[m]       \\ 
\cmidrule(r){1-8}
Camera 1                 & 1600×1200  & 16mm×12mm        & 45° & Sequence & Irregular~ & {[}0°, 70°] & [200, 500]  \\ 
Camera 2                 & 1920×1080  & 23.76mm×13.365mm & 25° & Grid       & Uniform    & 0°, 45° & [200, 500]      \\
Camera 3                 & 1920×1080  & 23.76mm×13.365mm & 25° & Sequence & Regular~   & 0°, 45° & [400, 500]      \\
\toprule
\end{tabular}
\caption{Camera configurations and sampling methodologies used for data acquisition.}
\label{tab:camera_configs}
\end{table*}


\section{Dataset}
\label{sec:Dataset}

\noindent We construct the \textbf{InsLoD-Loc} dataset, a large-scale synthetic dataset for building instance segmentation and 6-DoF localization. It comprises 108,109 RGB images with pixel-accurate instance annotations, covering 40 UAV flight areas across six countries (Japan, Switzerland, China, France, Italy, Netherlands).~\cref{fig:data} shows the geographic distribution of these areas. This section details the construction pipeline and composition of the \textbf{InsLoD-Loc} dataset.

\subsection{Simulation Environment and Data Generation}
\label{sec:Data_Generation}

\noindent \textbf{Photorealistic RGB Image Generation.}
To generate high-quality query images, we employed a comprehensive simulation framework based on UE5 version 5.1.1~\cite{UE_engine}. Urban scenes are built by streaming Google Photorealistic 3D Tileset data~\cite{GoogleEarth} via the Cesium for Unreal plugin~\cite{Cesium_GS}. We then utilized the integrated AirSim plugin from Microsoft Research version 2.1.0~\cite{Microsoft_Research} to render these scenes from multiple angles, at various altitudes, and across diverse locations. Further details are provided in the Appendix.

\noindent \textbf{Automated Instance Mask Generation.}
To generate instance masks with pixel-perfect alignment to the RGB images, we first source the corresponding geo-referenced LoD models~\cite{zhu2025lod, swisstopo, 3D_BAGS, PLATEAU, zamir2016generic}. These models are converted to the OBJ mesh format. We unify their coordinate systems to precisely match the reference system of the Google 3D Tileset data~\cite{GoogleEarth}. This enables us to render a corresponding instance mask via the real-time rendering engine OSG~\cite{OSG} using the \textit{exact same} camera intrinsic and extrinsic parameters as each rendered RGB frame. Finally, we apply Connected Component Analysis to these rendered instance maps to generate the precise instance mask annotations. Further details are provided in the Appendix.

\subsection{Data Acquisition and Composition}
\label{sec:Acquisition_and_Composition}

\noindent The dataset was collected from 40 geographically distinct areas using three camera configurations and the corresponding sampling methodologies to maximize diversity in viewpoint, altitude, and environment, as detailed in~\cref{tab:camera_configs}.

Specifically, the \textbf{InsLoD-Loc} dataset encompasses diverse land-use categories, including commercial, industrial, and residential, as well as education, medical, and suburban zones, to ensure comprehensive environmental representation. It is partitioned into geographically independent training, validation, and test sets. For completeness, we note that the RGB images for the China region are available in LoD-Loc v2~\cite{zhu2025lod}. Further details are provided in the Appendix.

\begin{table*}[htbp]
\centering
\begin{tabular}{cccccccccc} 
\toprule
\multicolumn{2}{c}{\multirow{2}{*}{Method}}                            & \multicolumn{4}{c}{\textit{in-Traj.}}                                 & \multicolumn{4}{c}{\textit{out-of-Traj.}}                               \\ 
\cmidrule(r){3-6} \cmidrule(r){7-10}
\multicolumn{2}{c}{}                                                   & 2m-2°          & 3m-3°          & 5m-5°          & T.e./R.e.          & 2m-2°          & 3m-3°          & 5m-5°           & T.e./R.e.           \\ 
\cmidrule(r){1-10}
\multicolumn{2}{c}{Prior}                                              & 0              & 0              & 4.30            & 6.48/1.63          & 0              & 0              & 0.36            & 11.10/0.92          \\ 
\hline
\multirow{5}{*}{CAD-Loc~\cite{panek2023visual}}                     & SIFT+NN                & 0              & 0              & 0              & -                  & 0              & 0              & 0               & -                   \\
                                              & SPP+SPG                & 0              & 0              & 0              & -                  & 0              & 0              & 0               & -                   \\
                                              & LoFTR                  & 0              & 0              & 0              & -                  & 0              & 0              & 0               & -                   \\
                                              & e-LoFTR                & 0              & 0              & 0              & -                  & 0              & 0              & 0               & -                   \\
                                              & RoMA                   & 0              & 0              & 0              & -                  & 0              & 0              & 0               & -                   \\ 
\hline
\multirow{2}{*}{MC-Loc~\cite{trivigno2024unreasonable}}                      & DINOv2                 & 1.20           & 4.10           & 17.40          & 8.29/2.58          & 2.40           & 7.40           & 26.10           & 7.02/2.29           \\
                                              & RoMa                   & 0.10           & 0.60           & 3.30           & 10.6/8.60          & 0.20           & 0.90           & 3.30            & 16.9/3.88           \\ 
\hline
LoD-Loc\dag~\cite{zhu2024lod}                                        & -                      & 49.56          & 71.82          & 89.09          & 3.32/1.48          & 54.20          & 75.05          & 89.51           & 3.33/1.18           \\ 
\hline
\multirow{3}{*}{LoD-Loc v2\dag~\cite{zhu2025lod}}                  & no refine              & 0              & 0              & 23.38          & 6.19/0.67          & 11.68          & 29.88          & 51.14           & 4.78/0.92           \\
                                              & no select              & 93.50          & 98.40          & 99.50          & 0.74/0.17          & 90.50          & 94.80          & 96.90           & 0.77/0.16           \\
                                              & Full                   & 93.70          & 98.40          & 99.50          & 0.72/0.15          & \textbf{97.90} & \textbf{99.80} & \textbf{100.00} & 0.71/0.14  \\ 
\hline
\multirow{3}{*}{\textbf{\textbf{LoD-Loc v3}}} & no refine              & 0              & 0              & 24.10          & 6.09/0.67          & 11.70          & 30.80          & 52.00           & 4.58/0.91           \\
                                              & no select              & 97.40          & 99.10          & 99.80          & 0.50/0.12          & 96.10          & 97.80          & 98.30           & 0.61/0.12           \\
                                              & \textbf{\textbf{Full}} & \textbf{97.60} & \textbf{98.90} & \textbf{99.70} & \textbf{0.49/0.13} & 97.40          & 99.00          & 99.40           & \textbf{0.60/0.12}           \\
\toprule
\end{tabular}
\caption{\textbf{Quantitative comparison results of different methods over UAVD4L-LoDv2 dataset.} T.e. and R.e. denote median translation error (m) and median rotation error (°). \dag~indicates models trained in-distribution on this dataset. Our method utilizes area-based weighting.}
\label{Tab:uavd4l}
\end{table*}

\begin{table*}[htbp]
\centering
\begin{tabular}{cccccccccc} 
\toprule
\multicolumn{2}{c}{\multirow{2}{*}{Method}}                    & \multicolumn{4}{c}{\textit{in-Place.}}                                      & \multicolumn{4}{c}{\textit{out-of-Place.}}                                   \\ 
\cmidrule(r){3-6} \cmidrule(r){7-10}
\multicolumn{2}{c}{}                                           & 2m-2°            & 3m-3°            & 5m-5°            & T.e./R.e.          & 2m-2°            & 3m-3°            & 5m-5°            & T.e./R.e.           \\ 
\cmidrule(r){1-10}
\multicolumn{2}{c}{Prior}                                      & 0                & 0                & 0.56             & 17.6/3.87          & 0                & 0                & 1.06             & 17.9/3.94           \\ 
\hline
CAD-Loc                                      & \textit{same*} & $\boldsymbol{0}$ & $\boldsymbol{0}$ & $\boldsymbol{0}$ & \textbf{-}                  & $\boldsymbol{0}$ & $\boldsymbol{0}$ & $\boldsymbol{0}$ & \textbf{-}                   \\ 
\hline
\multirow{2}{*}{MC-Loc}                      & DINOv2         & 0.90             & 4.40             & 17.50            & 8.18/2.54          & 2.90             & 9.00             & 30.20            & 6.23/1.97           \\
                                              & RoMa           & 0.20             & 1.20             & 4.80             & 9.80/2.65          & 0.70             & 2.10             & 11.5             & 10.3/3.97           \\ 
\hline
LoD-Loc\dag                                      & -              & 36.79            & 50.56            & 69.77            & 2.87/1.78          & 14.24            & 31.39            & 59.89            & 8.73/2.78           \\ 
\hline
\multirow{3}{*}{LoD-Loc v2\dag}                  & no refine      & 0.56             & 3.73             & 20.79            & 7.37/3.76          & 0.53             & 2.11             & 11.35            & 8.92/3.90           \\
                                              & no select      & 52.10            & 72.10            & 88.30
            & 1.90/0.89          & 31.10            & 55.90            & 81.30            & 2.73/0.73           \\
                                              & Full  & 54.20   & 74.60   & 92.00   & 1.83/0.85 & 31.40            & 58.53            & 86.30            & 2.64/0.73           \\ 
\hline
\multirow{3}{*}{\textbf{\textbf{LoD-Loc v3}}} & no refine      & 0.60             & 2.90             & 16.20            & 8.20/3.79          & 0.50             & 1.80             & 11.10            & 9.02/3.90           \\
                                              & no select      & 58.60            & 77.50            & 90.60            & 1.68/0.80          & 34.60            & 57.50            & 80.20            & 2.65/0.88           \\
                                              & \textbf{\textbf{Full}}           & \textbf{58.60}            & \textbf{79.90}            & \textbf{95.40}            & \textbf{1.61/0.77}          & \textbf{36.90}   & \textbf{64.10}   & \textbf{88.90}   & \textbf{2.43/0.83}  \\
\toprule
\end{tabular}
\caption{\textbf{Quantitative comparison results of different methods over Swiss-EPFLv2 dataset.} The \textit{same*} indicates that the statistics are identical to those in \cref{Tab:uavd4l}. \dag, T.e., and R.e. have the same meanings as those in \cref{Tab:uavd4l}. Our method utilizes area-based weighting.}
\label{Tab:table_swiss}
\end{table*}

\begin{table*}[htbp]
\centering
\begin{tabular}{cccccccccc} 
\toprule
\multicolumn{2}{c}{\multirow{2}{*}{Method}}                            & \multicolumn{4}{c}{\textit{Grid-Traj.}}                                     & \multicolumn{4}{c}{\textit{Sequence-Traj.}}                                  \\ 
\cmidrule(r){3-6} \cmidrule(r){7-10}
\multicolumn{2}{c}{}                                                   & 2m-2°            & 3m-3°            & 5m-5°            & T.e./R.e.          & 2m-2°            & 3m-3°            & 5m-5°            & T.e./R.e.           \\ 
\cmidrule(r){1-10}
\multicolumn{2}{c}{Prior}                                              & 0.10             & 1.70             & 8.90             & 8.03/1.78          & 0.30             & 0.80             & 8.80             & 8.19//1.76          \\ 
\hline
CAD-Loc                                           & \textit{same*}     & $\boldsymbol{0}$ & $\boldsymbol{0}$ & $\boldsymbol{0}$ & \textbf{-}         & $\boldsymbol{0}$ & $\boldsymbol{0}$ & $\boldsymbol{0}$ & \textbf{-}          \\ 
\hline
\multirow{2}{*}{MC-Loc}                           & DINOv2             & 0                & 0.26             & 1.61             & 31.95/1.88         & 3.42                & 10.50                & 29.81              & 5.16/0.49             \\
                                                  & RoMa               & 0                & 0.14             & 0.39             & 32.36/1.89         & 3.58                & 10.72                & 31.12              & 5.16/0.48           \\ 
\hline
LoD-Loc~                                          & -                  & 10.52            & 20.95            & 33.80            & 8.03/1.78          & 0                & 0                & 5.06             & 8.19/1.76           \\ 
\hline
\multirow{3}{*}{LoD-Loc v2~}                      & no refine          & 4.00             & 12.50            & 39.70            & 5.91/1.15          & 9.00             & 23.10            & 57.40            & 4.56/0.91           \\
                                                  & no select          & 2.50             & 7.50             & 23.50            & 8.04/1.56          & 2.90             & 7.30             & 24.60            & 8.53/1.43           \\
                                                  & Full               & 2.70             & 8.10             & 22.70            & 7.86/1.48          & 2.30             & 6.80             & 24.00            & 8.75/1.52           \\ 
\hline
\multirow{3}{*}{\textbf{LoD-Loc v3$_c$}}          & no refine          & 7.90             & 21.60            & 51.70            & 4.89/0.98          & 18.60            & 38.20            & 75.60            & 3.58/0.72           \\
                                                  &  no select & 35.00            & 62.00            & 83.40            & 2.53/0.30          & 41.60            & 69.10            & 89.80            & 2.24/0.28           \\
                                                  & \textbf{Full}      & \textbf{39.30}   & \textbf{68.00}   & \textbf{89.90}   & \textbf{2.29/0.27} & \textbf{50.30}   & \textbf{79.90}   & \textbf{97.30}   & \textbf{1.98/0.23}  \\ 
\hline
\multirow{3}{*}{\textbf{\textbf{LoD-Loc v3$_a$}}} & no refine          & 7.90             & 21.60            & 51.70            & 4.89/0.98          & 18.60            & 38.20            & 75.60            & 3.58/0.72           \\
                                                  &  no select &  35.50   &  61.70   &  85.10   &  2.49/0.29 &  42.30   &  71.70   &  92.70   &  2.21/0.27  \\
                                                  & \textbf{Full}      & \textbf{38.10}            & \textbf{65.40}            & \textbf{86.40}            & \textbf{2.42/0.27}          & \textbf{49.80}            & \textbf{79.90}            & \textbf{95.80}            & \textbf{2.00/0.23}           \\
\toprule
\end{tabular}
\caption{\textbf{Quantitative comparison results of different methods over Tokyo-LoDv3 dataset.} The \textit{same*} indicates that the statistics are identical to those in \cref{Tab:uavd4l}. T.e. and R.e. have the same meanings as in \cref{Tab:uavd4l}. LoD-Loc v3$_c$ method utilizes confidence-based weighting, and LoD-Loc v3$_a$ method utilizes area-based weighting.}
\label{Tab:table_tokyo}
\end{table*}
\begin{figure*}[htbp]
    \centering
    \includegraphics[ width=1.0\linewidth]{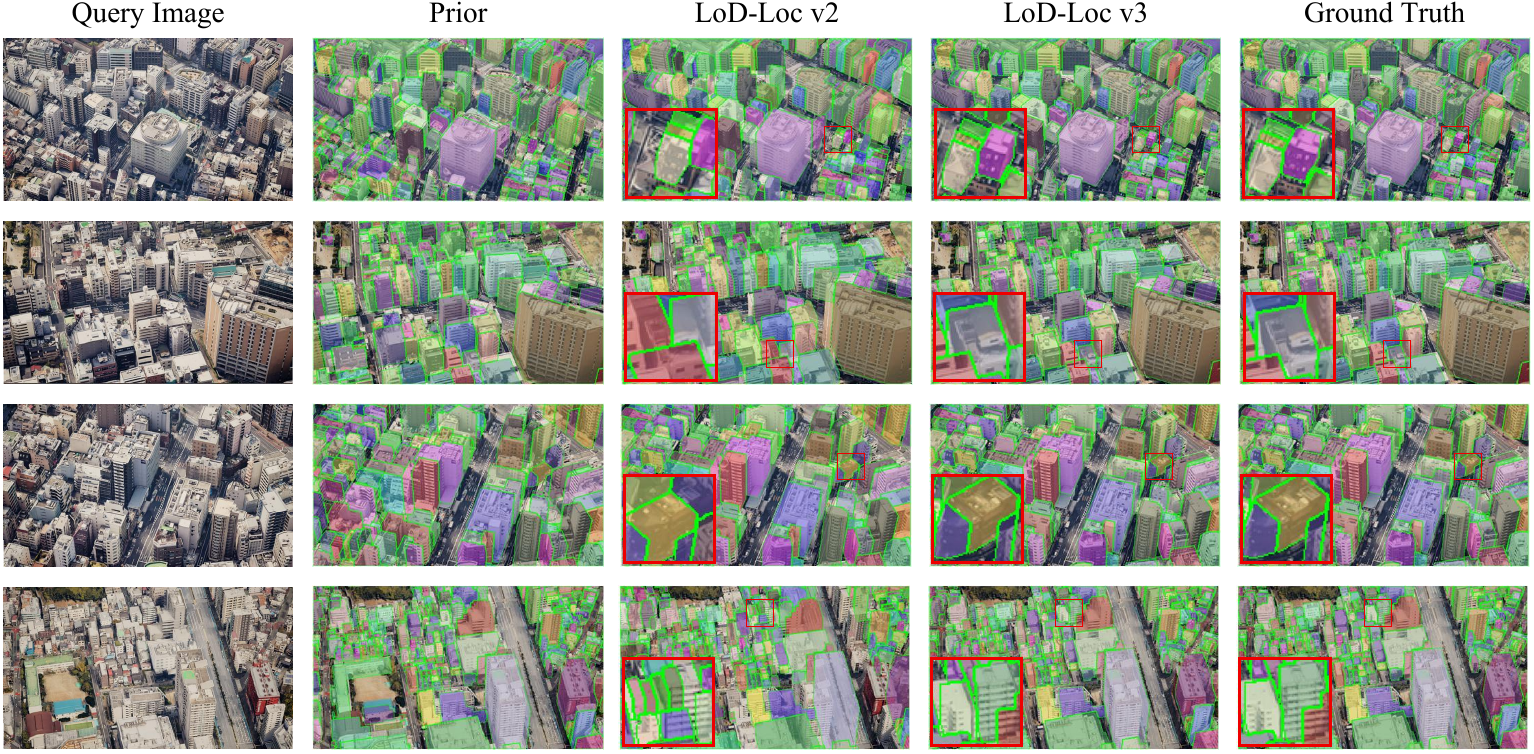}
    \caption{\textbf{Visualization of localization results on the Tokyo-LoDv3 dataset.} Superimposed instance masks rendered at estimated poses demonstrate that our method effectively resolves ambiguity in dense urban scenes. The columns from left to right show: query image, prior pose, LoD-Loc v2, LoD-Loc v3 (Ours), and ground truth.}
    \label{fig:exp_loc}
\end{figure*}

\begin{table*}[htbp]
\centering
\begin{tabular}{cccccccccc} 
\toprule
\multirow{2}{*}{Dataset}      & \multirow{2}{*}{Method}      & \multicolumn{4}{c}{\textit{in~(Grid)-Traj.~(Place.)}}           & \multicolumn{4}{c}{\textit{out-of~(Sequence)-Traj.~(Place.)}}     \\ 
\cmidrule(r){3-6} \cmidrule(r){7-10}
                              &                              & 2m-2$^\circ$   & 3m-3°          & 5m-5°          & T.e./R.e.          & 2m-2°          & 3m-3°          & 5m-5°          & T.e./R.e.           \\ 
\cmidrule(r){1-10}
\multirow{2}{*}{UAVD4L-LoDv2} & LoD-Loc v2                   & 71.40          & 91.20          & 96.10          & 1.66/0.34          & 70.00          & 89.80          & 96.10          & 1.62/0.39           \\
                              & \textbf{LoD-Loc v3}          & \textbf{97.60} & \textbf{98.90} & \textbf{99.70} & \textbf{0.49/0.13} & \textbf{97.40} & \textbf{99.00} & \textbf{99.40} & \textbf{0.60/0.12}  \\ 
\hline
\multirow{2}{*}{Swiss-EPFLv2} & LoD-Loc v2                   & 11.90          & 28.10          & 52.00          & 4.79/1.42          & 24.00          & 49.30          & 77.80          & 3.02/1.12           \\
                              & \textbf{LoD-Loc v3}          & \textbf{58.60} & \textbf{79.90} & \textbf{95.40} & \textbf{1.61/0.77} & \textbf{36.90} & \textbf{64.10} & \textbf{88.90} & \textbf{2.43/0.83}  \\ 
\hline
\multirow{2}{*}{Tokyo-LoDv3}  & LoD-Loc v2                   & 22.70          & 43.80          & 74.70          & 2.91/0.32          & 35.60          & 65.70          & 92.00          & 2.41/0.27           \\
                              & \textbf{\textbf{LoD-Loc v3}} & \textbf{39.30}          & \textbf{68.00}          & \textbf{89.90}          & \textbf{2.29/0.27}          & \textbf{50.30}          & \textbf{79.90}          & \textbf{97.30}          & \textbf{1.98/0.23}           \\
\toprule
\end{tabular}
\caption{\textbf{Ablation on Silhouette Representation.} Quantitative comparison of LoD-Loc v2 (semantic) and LoD-Loc v3 (instance) trained on the same InsLoD-Loc dataset.}
\label{Tab:ablation}
\end{table*}
\begin{table}[htbp]
\centering
\setlength{\tabcolsep}{4pt} 
\begin{tabular}{ccc} 
\toprule
\multirow{2}{*}{Alignment} & $Grid$-$Traj.$             & $Sequence$-$Traj$           \\ 
\cmidrule(r){2-3}
                           & 2m-2°/3m-3°/3m-3° & 2m-2°/3m-3°/3m-3°  \\ 
\hline
LoD-Loc v2                 & 19.60/39.40/72.10          & 21.50/47.80/89.00           \\
LoD-Loc v3                 & 38.10/65.40/86.40          & 49.80/79.90/95.80           \\
\toprule
\end{tabular}
\caption{\textbf{Ablation on Instance Alignment.} Quantitative comparison of localization performance using separated instance masks versus artificially merged semantic masks.}
\label{tab:instance_alignment}
\end{table}
\section{Experiments}
\label{sec:Experiments}

\noindent In this section, we first introduce datasets and baseline methods,
as well as evaluation metrics in~\cref{sec:Experiment_Settings_and_Baselines}, followed by implementation details of our method in~\cref{sec:Implementation_Details}.
Experimental results and ablation studies are detailed in~\cref{sec:Evaluation_Results} and \cref{exp:ab_study}, respectively.

\subsection{Experiment Settings and Baselines}
\label{sec:Experiment_Settings_and_Baselines}

\noindent \textbf{Datasets.}
\noindent We evaluate our method on three datasets: UAVD4L-LoDv2, Swiss-EPFLv2, and Tokyo-LoDv3. The UAVD4L-LoDv2 and Swiss-EPFLv2 datasets cover areas of 2.5~km$^2$ and 8.2~km$^2$. Details of these two public benchmarks can be found in LoD-Loc v2~\cite{zhu2025lod}. Tokyo-LoDv3 serves as the test set from the \textbf{InsLoD-Loc} dataset. It covers six land-use categories and features five dense urban scenes along with corresponding instanced LoD1 models. Further details are provided in the Appendix.

\noindent \textbf{Baselines.}
We compare against three baseline categories: those that perform localization via feature matching, namely CAD-Loc~\cite{panek2023visual}; those that perform localization via feature alignment,  namely MC-Loc~\cite{trivigno2024unreasonable}; and those that operate by aligning geometric primitives from LoD models, namely LoD-Loc~\cite{zhu2024lod} and LoD-Loc v2~\cite{zhu2025lod}. Specifically, for the feature-matching category, we evaluate CAD-Loc with a suite of keypoint-based strategies: 1) the classic SIFT~\cite{lowe2004distinctive} descriptor with Nearest Neighbor (NN) matching; 2) the learning-based SuperPoint (SPP)~\cite{detone2018superpoint} extractor with the SuperGlue (SPG)~\cite{sarlin2020superglue} matcher; 3) the detector-free matcher LoFTR~\cite{sun2021loftr} and its efficient variant 4) e-LoFTR~\cite{wang2024efficient}; and 5) the dense feature matcher RoMa~\cite{edstedt2024roma}. For the feature-alignment category, we evaluate MC-Loc using two distinct feature extraction backbones: 6) a pre-trained DINOv2~\cite{oquab2023dinov2} encoder and 7) the RoMa~\cite{edstedt2024roma} dense feature extractor. For the LoD-based alignment category, 8) LoD-Loc~\cite{zhu2024lod} is implemented with its default parameters, and 9) we evaluate the three reported variants of LoD-Loc v2~\cite{zhu2025lod}. Implementation details for the baseline experiments are provided in the Appendix.

\noindent \textbf{Evaluation Metrics.}
Following standard localization evaluation protocols~\cite{toft2020long}, we report the percentage of queries localized within $(2\text{m}, 2^{\circ})$, $(3\text{m}, 3^{\circ})$, and $(5\text{m}, 5^{\circ})$ thresholds, consistent with LoD-Loc~\cite{zhu2024lod}.

\subsection{Implementation Details}
\label{sec:Implementation_Details}

\noindent \textbf{Training.} Our segmentation module is trained on the 88,493-image training split of \textbf{InsLoD-Loc}. The input image size is set to \( (512, 512) \). To optimize the training process, we use the AdamW optimizer with a base learning rate of \( 2 \times 10^{-4} \) and a weight decay of $0.05$. We employ a cosine annealing schedule and train for $20$ epochs. The SAM encoder is based on the pre-trained \texttt{ViT-Huge} model and is fine-tuned using the parameter-efficient LoRA technique.

\noindent \textbf{Inference.} Inference is performed on test sets differing in region and viewpoint from the training set. Input size remains (512, 512). For the coarse pose selection stage, the query and rendered candidate mask sizes are adapted to each dataset (e.g., $640 \times 360$) with a 10m sampling interval. In the fine pose estimation stage, the mask sizes are identical. Consistent with LoD-Loc v2~\cite{zhu2025lod}, we set the number of iterations to $40$, using 2 beams and an angular perturbation of $2^{\circ}$. The translational perturbation follows a Gaussian distribution \( X \sim \mathcal{N}(0, \sigma^2) \) with a mean of 0 and \( \sigma = 1.5 \). The decay factor is set to \( \gamma = 0.3 \), and \( n = 52 \) candidates are generated per iteration. All training and testing were performed on two NVIDIA RTX 4090 GPUs.

\subsection{Evaluation Results}
\label{sec:Evaluation_Results}

In this section, we compare our results with SOTA visual localization frameworks to validate the superiority of our approach. We also provide visualizations to highlight key aspects and present ablation studies.

\noindent \textbf{Evaluation over UAVD4L-LoDv2 dataset.}
As shown in~\cref{Tab:uavd4l}, our method performs excellently in both \textit{in-Traj.} and \textit{out-of-Traj.} queries, achieving highly competitive performance compared to LoD-Loc v2~\cite{zhu2025lod}. Crucially, LoD-Loc v2 was trained in-distribution, whereas LoD-Loc v3 was trained *only* on our synthetic \textbf{InsLoD-Loc} dataset, demonstrating strong cross-domain generalization.

\noindent \textbf{Evaluation over Swiss-EPFLv2 dataset.}
Results on Swiss-EPFLv2, as shown in~\cref{Tab:table_swiss}, further confirm this. Our method significantly surpasses all baselines, including the in-distribution-trained LoD-Loc v2, across all metrics, despite our model relying solely on synthetic data.

\noindent \textbf{Evaluation over Tokyo-LoDv3 dataset.}
The Tokyo-LoDv3 dataset features five challenging dense urban scenes. As shown in~\cref{Tab:table_tokyo}, LoD-Loc v3 achieves SOTA accuracy in both \textit{Grid-Traj.} and \textit{Sequence-Traj.} queries. In contrast, LoD-Loc v2, which relies on semantic segmentation, fails completely in these scenes due to ambiguity. This validates our instance-alignment strategy for dense scenes, as qualitatively demonstrated in~\cref{fig:exp_loc}. We note a slight misalignment in the source LoD models for this region, which affects absolute accuracy but not the comparative result.

\subsection{Ablation Study}
\label{exp:ab_study}

We conducted ablation studies on the UAVD4L-LoDv2, Swiss-EPFLv2, and Tokyo-LoDv3 datasets, focusing on the silhouette representation and alignment methods.

\noindent \textbf{Silhouette Representation.}
To isolate the performance gains brought by our instance-level representation from the benefits of the new dataset, we retrained the LoD-Loc v2 semantic segmentation model on our \textbf{InsLoD-Loc} dataset. Evaluated on the three aforementioned datasets, the retrained LoD-Loc v2 remains significantly inferior to LoD-Loc v3, as shown in~\cref{Tab:ablation}. This persistent performance gap confirms that our superior generalization stems fundamentally from the paradigm shift to instance-level representation, rather than merely from scaling up the training data.

\noindent \textbf{Instance Alignment.}
To explicitly validate the efficacy of the instance alignment mechanism, we conducted an ablation where our predicted instance masks were forcefully merged into a single semantic mask during the pose evaluation stage. As shown in~\cref{tab:instance_alignment}, we validate the superiority of instance masks in Tokyo-LoDv3 datasets, demonstrating that instance alignment is crucial for resolving semantic ambiguities. Further details on the visualization of results and additional ablation studies are provided in the Appendix.
\section{Conclusion}
\label{sec:Conclusion}

\noindent This paper presents LoD-Loc v3, a novel aerial localization framework that effectively tackles the challenges of cross-scene generalization and localization ambiguity in dense urban environments. Our solution is twofold: a large-scale synthetic dataset, InsLoD-Loc, enables the learning of domain-invariant features enhancing model generalization, while a novel paradigm, shifting from semantic to instance-level silhouette alignment, effectively resolves ambiguities caused by merged structures in dense cities. Extensive experiments demonstrate that our method achieves SOTA cross-domain performance and succeeds in complex scenes where prior methods fail, significantly advancing the practicality of global-scale aerial localization.

\noindent \textbf{Limitations.} Our method's performance is partially dependent on the accuracy of the instance segmentation. Segmentation failures, particularly under extreme adverse weather conditions, may lead to degraded localization accuracy. Further details on the localization results under extreme weather are provided in the Appendix.
{
    \small
    \section*{Acknowledgments}
    The authors would like to acknowledge the support from the National Natural Science Foundation of China, Grant No. 62406331.
    \bibliographystyle{ieeenat_fullname}
    \bibliography{main}
}


\end{document}